 \documentclass[mlmain]{jmlr}

\newcommand{\R}{\bb{R}}

\newcommand{\dd}{\,\mathrm{d}}

\newcommand{\Nat}{\bb{N}}

\newcommand{\ip}[2]{\langle #1 , #2 \rangle}

\newcommand{\mc}[1]{\mathcal{#1}}
\newcommand{\bb}[1]{\mathbb{#1}}

\newcommand{\spanv}{\operatorname{span}}

\newcommand{\CL}{\mc{C}_L}

\renewcommand{\set}[1]{\left\{#1\right\}}
\newcommand{\norm}[1]{\left\Vert #1 \right\Vert}

\usepackage{longtable}% for long tables
\usepackage{booktabs}
\usepackage[load-configurations=version-1]{siunitx} % newer version
\usepackage{float} % for [H] float placement

\theorembodyfont{\upshape}
\theoremheaderfont{\scshape}
\theorempostheader{:}
\theoremsep{\newline}

\jmlrvolume{}
\firstpageno{1}
\jmlryear{2026}
\jmlrworkshop{Symmetry and Geometry in Neural Representations}
\title[Composition Operators for LLM Transformations]{Using Composition Operators to Linearize \\ LLM Semantic Transformations}

  \author{\Name{Afjal Chowdhury} \Email{afjal@mit.edu}\\
  \Name{James Chen} \Email{jchen163@mit.edu}\\
  \Name{Alan Edelman} \Email{edelman@mit.edu}\\
  \addr Massachusetts Institute of Technology}

\begin{document}

\maketitle

\begin{abstract}
Machine learning learns functions: prompt to response, image to caption. What these functions are mathematically remains hard to say. We present a method to approximate these kinds of transformations using techniques from dynamical systems that fall under the umbrella of Koopmanism. We introduce the use of composition operators, which generalize the Koopman operator and, crucially, can map between distinct spaces, motivating the perspective that LLM transformations are rectangular infinite-dimensional operators. This formalism reveals useful structure: under natural assumptions on the prompt and response distributions, the LLM operator is an isometry, and misalignment between learned representations manifests as spectral pollution of its finite sections. We then outline a method of constructing finite-dimensional approximations of an LLM operator, and demonstrate how the singular value spectrum can be used to compare tasks and models.
\end{abstract}

\section{Introduction}
\label{sec:intro}
A Large Language Model (LLM), at least in the deterministic case, can be thought of as a mathematical function $L : \mathcal{X} \rightarrow \mathcal{Y}$ between prompts $\mc{X}$ and responses $\mc{Y}$. A core question is whether this function has any useful mathematical properties and to what extent we are able to measure them. These questions are particularly important from an interpretability perspective, especially for comparing different model behaviors \citep{Klabunde2025Similarity}, and for understanding composition in systems with multiple interacting agents \citep{scofield2026multiagent}.

The main difficulty is that LLMs are composed of highly nonlinear functions, which are challenging to analyze. Techniques from infinite-dimensional machine learning \citep{boulle2024mathematical,scholkopf2002learning} allow us to make sense of such functions by ``lifting'' objects from the domain into an infinite-dimensional space of observables, which can be thought of as the space of all features. Much of the work in infinite-dimensional machine learning centers around Koopman operator theory. Given a (typically nonlinear) discrete dynamical system on a state space $\mc{X}$ described by a transition map $F$, Koopman theory associates a \emph{linear} operator $\mathcal{K}_{F}$ on the space of observables on $\mc{X}$, acting on observables by composition with $F$:
\begin{equation}
(\mathcal{K}_{F}\phi)(x) = \phi(F(x)).
\end{equation}
Thus, one trades nonlinearity on a smaller state space for linearity on an infinite-dimensional space. Many methods exist to approximate the Koopman operator $\mathcal{K}_{F}$, notably the family of Dynamic Mode Decomposition (DMD) algorithms \citep{colbrook2023multiversedynamicmodedecomposition}. The simplest methods take a finite section of the infinite-dimensional operator $\mathcal{K}_{F}$: given a subspace of observables $V$ of dimension $n$, the matrix representing $\mc{P}_V \mc{K}_F|_V$ is $n \times n$ and describes the action of $\mc{K}_F$ restricted to $V$.

Koopman theory enforces that the domains and codomains are the same. Our work departs from this framework by considering the broader class of \emph{composition operators}, which are no longer constrained to be induced by a self-map $F$. This generalization enables functions that map between distinct spaces, to be analyzed using analogous techniques:
\begin{equation}
(\CL \phi) (x) = \phi(L(x)).
\end{equation}
Such composition operators are classical in analysis \citep{singh1993composition}, and have also previously appeared as functional maps between shapes in \citep{ovsjanikov2012functional} and \citep{glashoff2017compositionoperatorsmatrixrepresentation}, where the authors in the latter also consider the finite section method for computing functional maps.

Many machine learning problems have asymmetric inputs and outputs, e.g., image generation, making Petrov--Galerkin rectangular approximations more appropriate than Extended Dynamic Mode Decomposition (EDMD) \citep{williams2015edmd}. From the Koopman perspective, rectangular truncations of square operators have been considered in \citep{colbrook2026prone}. Restricting the domain and codomain to chosen finite-dimensional observable spaces yields a finite-dimensional linear approximation of the underlying operator.

Lastly, Koopman operators convert composition of maps into operator multiplication. More precisely, $\mc{K}_G \mc{K}_F = \mc{K}_{F \circ G}$, a property maintained by general composition operators and is approximately preserved under truncations, enabling the study of LLM composition and agent interactions.

\noindent Our main contributions are as follows:

\noindent \textbf{Composition Operator Viewpoint.} We motivate and derive a Koopman-inspired framework for general composition operators, retaining the numerical methods that make Koopman analysis so powerful while dropping superfluous structure, allowing analysis of mappings that do not make sense as flows, like sensory-motor mappings in neuroscience \citep{lam2026sensory}. We further find that, under assumptions outlined in Section \ref{sec:cds}, LLM operators are isometries, warranting new interpretations of the finite section method. 

\noindent \textbf{Embeddings as Subspaces of Feature Space.} Features, of the form $\mc{X} \rightarrow\R$, are analogous to Koopman observables, so embedding maps induce finite-dimensional subspaces of an infinite-dimensional feature space (Section \ref{f-as-f-space}), which we relate to the prompt and response distributions in Section \ref{sec:cds}. 

\noindent \textbf{Reinterpreting Regression of Embeddings.}
By viewing embedding space as approximate subspace of infinite-dimensional feature space, regression on embeddings can be reinterpreted as a numerical spectral method on a contextual composition operator (Section \ref{sec:fs-cco}). We demonstrate in Section \ref{sec:RSA} that RSA techniques like CCA are finite section calculations. 

\noindent \textbf{Spectral Comparison of LLM Transformations.} In Section \ref{sec:comparing}, we compute finite sections of LLM operators on prompt--response data, and demonstrate how tasks of increasing open-endedness exhibit systematically different singular value spectra, enabling quantitative comparison of tasks. We also demonstrate that the finite sections capture statistically significant information of the operators.

\section{Koopmanism}
Despite our departure from using Koopman operators, techniques that are ubiquitous in Koopman analysis remain crucial to our derivation; in this section we detach them from the language of dynamical systems so that they can be reinterpreted in our framework.
\subsection{Features as Observables}
\label{f-as-f-space}
Learning vector representations of data $\mc{X}$ is a central task in machine learning \citep{bengio2013representation} and neuroscience \citep{chung2021neural}. Representations are functions of the form  $\Phi : \mathcal{X} \rightarrow \R^d$ that assign an input in some discrete data domain $\mathcal{X}$ (e.g., words or images) to a set of $d$ features. The size of a representation is somewhat arbitrary: \citep{mezic2023operator} considers indicator feature maps whose entries $\phi_i$ indicate the presence of $d$ different entities within an image; one can consider new entities ad infinitum as $d \rightarrow \infty$. For $B = \{\phi_i\}_{i\in \Nat}$, consider the pseudo-representation 
\begin{equation}
\Phi_B: \mathcal{X} \rightarrow \ell^2(\Nat), \quad\Phi_B(x) = [\phi_1(x), \phi_2(x), \phi_3(x), \dots]^T.
\end{equation}
We can understand these kinds of infinite representations through the lens of \emph{lifting} in Koopman theory, where states in $\mc{X}$ are mapped into a separable Hilbert space $\mc{F}(\mc{X})$ \citep{koopman1931hamiltonian} like $L^2(\mc{X})$ or an RKHS \citep{das2020koopman}. Here $\mathcal{F}(\mathcal{X})$ is the space of all features or observables of the form $f:\mathcal{X} \rightarrow \R$. We later take this to be a \emph{contextual feature space}, which we define in Section \ref{sec:cds}.

Hilbert spaces are equipped with an inner product and, via Gram--Schmidt, a means of constructing orthonormal bases $B$, which for $\Phi_B$ corresponds to well chosen  entities so that the $\phi_i$ capture all of the semantics. Fixing $B$ allows any bounded linear operator $\mathcal{A} : \mc{F}(\mc{X}) \rightarrow \mc{F}(\mc{X})$ to admit an infinite matrix representation $([\mc{A}]_B)_{ij} = \langle\phi_i, \mc{A}\phi_j \rangle$. %Throughout, $\cdot^T$ denotes the transpose of a real matrix (finite or infinite), while $\cdot^*$ is reserved for the adjoint of an operator between Hilbert spaces; since all matrices in this work are real, the matrix representing an adjoint is the transpose of the original matrix representation.

\subsection{Finite Section Approximations}
% Recent works on Koopman Autoencoders (KAEs)  \citep{aswani2025koopman, berman2023multifactor, otto2019linearly} uncover dynamical properties of representations evolving across the layers of a deep network using EDMD.

Bounded linear operators on separable Hilbert spaces admit a spectrum, which is useful information when considering LLMs as operators, but its calculation is difficult when working with infinite matrices. We instead solve these problems numerically using the finite section method \citep{hansen2008approximation, arveson1994cstar}. These are also known as Galerkin approximations and are used in methods like EDMD for Koopman operators. The simplest example of the finite section method on an operator $\mc{C}$ is to simply use the top-left $n \times n$ submatrix of the infinite matrix $[\mc{C}]_B$. This is equivalent to taking the first $n$ basis vectors of $\Phi_B$ as a \emph{dictionary}, called $\Phi_n$, and constructing:
\begin{equation}
\label{fs}
   K_n = \mathcal{P}_{\Phi_n} \mathcal{C} \mathcal{P}_{\Phi_n}^* \in \R^{n\times n}.
\end{equation}
Here, $\mc{P}_\Phi : \mc{F}(\mc{X}) \rightarrow \R^n$ is the \emph{analysis operator} of the dictionary $\Phi$, $(\mc{P}_\Phi f)_i = \ip{\phi_i}{f}$, whose adjoint $\mc{P}^*_\Phi$ is the synthesis operator $c \mapsto c^T\Phi$ \citep{lusch2018deep}. When the dictionary is orthonormal, $\mc{P}_\Phi f$ is exactly the vector of expansion coefficients of the projection of $f$ onto $V_\Phi$ (for a general dictionary the coefficient map is instead $G_\Phi^{-1}\mc{P}_\Phi$, with $G_\Phi$ the Gram matrix of Section \ref{sec:svals}), and $\mc{P}_\Phi$ satisfies:
\begin{equation}
\label{proj}
    \mc{P}_\Phi \mc{P}^*_\Phi = I_n, \quad\text{and}\quad\mc{P}^*_\Phi \mc{P}_\Phi = \Pi_{V_\Phi},
\end{equation}
where $\Pi_{V_\Phi}$ denotes the orthogonal projection onto $V_\Phi = \spanv\set{\phi_1,\dots,\phi_n}$.

One can also consider a rectangular section of an operator by considering two different dictionaries $\Phi_n$ and $\Psi_m$, in which case we approximate the singular values of the operator $\mc{C}$. This has recently been used to approximate Koopman operators in \citep{colbrook2026prone}, and in spectral methods for PDEs \citep{townsend2015automatic, shen2003dual}.

An important caveat with these methods is the issue of spectral pollution \citep{davies2004spectral}, where spurious eigenvalues (or singular values in the rectangular case) appear in the empirical spectrum, typically due to a poor choice of dictionary $\Phi$. Algorithms like ResDMD exist to mitigate this problem for Koopman operators \citep{colbrook2023residual, colbrook2024rigorous}.

\subsection{Example: Representational Similarity Analysis}
\label{sec:RSA}

Representational Similarity Analysis (RSA) is a tool used in machine learning and theoretical neuroscience to compare how the same stimuli $W$ (e.g., words or images) are structured under different representations $\Phi : W \rightarrow \R^n$ and $\Psi : W \rightarrow \R^m$ \citep{kornblith2019similarity, sucholutsky2025getting}. This classically boils down to a comparison of the high-dimensional point clouds induced by $\Phi, \Psi$ on the items in $W$.

We previously explained that finite-dimensional representations are approximate representations of a ground truth that is infinitely precise, so it makes sense to think of RSA tools that compare different representations as measuring an ``error'' of some kind. We demonstrate that this error is related to spectral pollution, which arises from a sub-optimal choice of representations in the finite section method, by establishing an equivalence between Canonical Correlation Analysis (CCA) \citep{hotelling1936relations}, a classic RSA method, and the finite section method as a numerical spectral calculation.

CCA identifies a sequence of $N \leq \min(n,m)$ maximally correlated $1$D projections between paired datasets, which each have a corresponding canonical correlation coefficient $1\geq \rho_1\geq \dots \geq\rho_N \geq 0$. The average correlation coefficient, or the average squared coefficient, is often used as a summary statistic of overall representational similarity. We demonstrate that these coefficients are precisely the empirical singular value spectrum of a finite section of a unitary operator, using whitened, mean-centered representations
\begin{equation}
\label{proj-whitened}
    \tilde\Phi = G_\Phi^{-\frac{1}{2}}\Phi, \quad \tilde\Psi =G_\Psi^{-\frac{1}{2}}\Psi; \qquad \mc{P}_{\tilde\Phi} = G_{\Phi}^{- \frac{1}{2} } \mc{P}_\Phi, \quad \mc{P}_{\tilde\Psi} = G_{\Psi}^{- \frac{1}{2} } \mc{P}_\Psi.
\end{equation}

We lift $W$ to a function space $\mc{F}(W)$ (as in Section \ref{f-as-f-space}) and consider the identity operator on this space, $\mathrm{id}_{\mc{F}(W)}$. By taking a Petrov--Galerkin rectangular section $K$ using the two representations $\tilde\Phi, \tilde\Psi$, we essentially ask whether these representations are able to capture an underlying operator which encodes ``being equivalent''. We can then compare the empirical singular value spectrum of $K$ to $\{1\}$, framing misalignment between representations as spectral pollution/leakage. Section \ref{sec:fs-cco} outlines the calculation for how, given paired data $(\Phi(w_i), \Psi(w_i))$, one can make a data-driven approximation of the section
\begin{equation}
K = \mathcal{P}_{\tilde\Phi}\, \mathrm{id}_{\mc{F}(W)}\, \mathcal{P}^*_{\tilde\Psi} \in \bb{R}^{n \times m}.
\end{equation}
Using identity \eqref{proj-whitened}, and defining $(\Sigma_{xy})_{ij} = [\mc{P}_\Phi \mc{P}^*_\Psi]_{ij} = \ip{\phi_i}{\psi_j}$ (so $G_\Phi = \Sigma_{xx}, G_\Psi = \Sigma_{yy}$), we recover the whitened cross-covariance matrix:

% btw i found out this rn
% the distribution of the null hypothesis eigenvalues is given by the equation 
% \frac{\sqrt{4-\frac{x}{b}(1-b)^{2}}}{2\pi\sqrt{\frac{x}{b}}(1+\frac{bx}{b})}\cdot\frac{1}{b}
% if you set b = 1/8 = r/n then take the quantile function i.e. F^{-1}(1-x) where F is the cdf of that distribution, then this curve matches *exactly* the curve in the Figure.

\begin{equation}
K = G_\Phi^{- \frac{1}{2}} \mc{P}_\Phi \mc{P}^*_\Psi G_\Psi^{-\frac{1}{2}} = \Sigma_{xx}^{- \frac{1}{2}} \Sigma_{xy} \Sigma_{yy}^{-\frac{1}{2}},
\end{equation}
whose singular values $\sigma(K)$ are precisely the canonical correlation coefficients $\rho_i$ \citep{hotelling1936relations}. Further to this, \citep{pmlr-v285-williams24a} outlines an equivalence between CCA and other popular RSA methods like CKA. This indicates that there is potential for a whole host of RSA metrics to be viewed as the same kind of numerical spectral method.

\section{Contextual Composition Operators}
\label{sec:contextual_comp_op}

We return to the analysis of composition operators \cite{singh1993composition}. Given a map $L:\mc{X \rightarrow Y}$ with discrete spaces $\mc{X}$ and $\mc{Y}$, we will consider composition operators of the form
\begin{equation}
    \CL: L^2(\mc{Y}, \nu) \to L^2(\mc{X}, \mu), \qquad \CL f(x) = f(Lx),
\end{equation}
where $\mu$ and $\nu$ are measures on $\mc{X}$ and $\mc{Y}$ respectively. In order for this to be well defined, we need $L_\sharp \mu \ll \nu$, where $L_\sharp \mu(A) = \mu(L^{-1} A)$ is the \emph{pushforward} measure of $\mu$.

Since the spaces are discrete, this condition can be rephrased as: $\nu(\set{y}) = 0$ implies that $\mu(\set{x}) = 0$ for any $x$ such that $Lx = y$. In particular, the image of the support of $\mu$ must be contained in the support of $\nu$.
Under this condition, the operator is always closable and the Radon--Nikodym derivative exists and is given by
\begin{equation}
    \frac{\dd L_\sharp \mu}{\dd \nu}(y) = \frac{\mu(L^{-1} \set{y})}{\nu(\set{y})}.
\end{equation}
One can then show that the operator $\CL$ is densely defined if and only if $\frac{\dd L_\sharp \mu}{\dd \nu}(y) < \infty$ for every $y \in \mc{Y}$ and that $\CL$ is continuous if and only if $\frac{\dd L_\sharp \mu}{\dd \nu} \in L^\infty(\mc{Y}, \nu)$. The requirements that $\CL$ is closable and densely defined ensure that the operator has an adjoint $\CL^*$.

\subsection{Contextual Distributions}
\label{sec:cds}

Specializing to the case of LLMs, we set $\mc{X}$ and $\mc{Y}$ to be the space of token sequences $\Sigma^*$ and impose that $\mu$ and $\nu$ are probability distributions, now denoted $\rho_x$ and $\rho_y$ and called the \emph{contextual distributions}: $\rho_x$ represents the distribution of possible inputs (e.g., English sentences for an LLM used for translation) and $\rho_y$ the distribution of possible outputs, both specialized to the task the map $L$ performs. Thus we can define \emph{contextual feature spaces} as spaces of the form $\mc{H}_x := L^2(\Sigma^*,\rho_x)$, so that $\CL : \mc{H}_y \rightarrow \mc{H}_x$. Composition operators of this form are called \emph{contextual composition operators}.

\begin{remark}
    Finiteness of the measure $\rho$ ensures that the space $L^2(\Sigma^*,\rho)$ is rich enough to contain all bounded functions $\phi:\Sigma^* \to \R$ (which are not square-integrable with respect to the usual counting measure on $\Sigma^*$).  This is useful in practice, since observables such as embedding coordinates and LLM hidden states are typically bounded.
\end{remark}
The pushforward of $\rho_x$ under $L$ can then be interpreted as the distribution on $\Sigma^*$ of all possible outputs of the LLM given a random starting prompt drawn from $\rho_x$. A natural choice for $\rho_y$ is simply to take the response measure to be exactly the pushforward of the prompt measure, i.e., $\rho_y = L_\sharp \rho_x$. This condition is equivalent to $\CL$ being an isometry (under the assumption that $L$ is deterministic): indeed, the Radon--Nikodym derivative is the constant function $1$ and thus $\CL^* \CL = I$. Taking $\rho_x, \rho_y$ to be the prompt and response distributions allows the interpretation that the LLM map $L$ pushes the prompt distribution $\rho_x$ forward to $\rho_y$.
%  \red{[This presumes $L$ is deterministic (e.g. greedy decoding) — worth stating here, not only in the Discussion, since a sampled LLM maps a prompt to a distribution and the pushforward argument changes.]} 
\subsection{Approximation of Finite Sections of Contextual Composition Operators}
\label{sec:fs-cco}
Under the assumptions of the previous section, $\CL$ is an isometry, so finite sections may seem uninteresting. However, most observables in $\mc{H}_x$ and $\mc{H}_y$ contain little useful information, motivating us to study $\CL$ restricted to observables that capture useful features, i.e., representations. This parallels the restriction of Koopman operators to selected observables in EDMD \citep{williams2015edmd}.

Given paired prompt and response data $(x_i, y_i)$ where $y_i = Lx_i$, and two learned (potentially whitened, cf.\ \eqref{proj-whitened}) representations $\Phi, \Psi$ associated with prompts and responses respectively, we wish to find the best matrix $K \in \R^{N \times M}$ which linearly predicts the response features $\Psi(Lx) \in \R^M$ from the prompt features $\Phi(x) \in \R^N$, i.e., a $K$ such that:
\begin{equation}
\label{eq:linear_approx}
    K^T\Phi(x)  \approx \Psi(Lx).
\end{equation}
A priori it is not clear why we should expect a linear relation to occur here, and hence why a good approximation $K$ should exist. However, we can write this in terms of $\CL$ to obtain
\begin{equation}
    \sum_{j=1}^N K_{ji} \phi_j \approx \CL \psi_i \qquad i=1,\dots,M ,
\end{equation}
from which we can see that $K$ is the section of $\CL$ restricted to the subspaces spanned by the entries of $\Phi$ and $\Psi$. In other words, $K$ represents $\Pi_{V_{\Phi}} \CL |_{V_{\Psi}}$ where $V_{\Phi}$ and $V_{\Psi}$ are the subspaces spanned by the corresponding observables. Thus, the representations $\Phi$ and $\Psi$ act as trial and test functions in a Petrov--Galerkin method \citep{colbrook2026prone}. Throughout, $K \in \R^{N\times M}$ denotes this section matrix (rows indexed by $\Phi$, columns by $\Psi$): $K = \mc{P}_{V_\Phi}\CL\mc{P}^*_{V_\Psi}$ for orthonormal dictionaries, and in general the coefficient matrix $G_\Phi^{-1}\mc{P}_{V_\Phi}\CL\mc{P}^*_{V_\Psi}$.

Thus, our goal is to compute this section of $\CL$ given the dictionaries of observables $\Phi$ and $\Psi$. In Appendix \ref{sec:minimiser} we define what it means to be the best approximator in our situation, in a similar fashion to the derivation of the finite sections of Koopman operators \citep{colbrook2023multiversedynamicmodedecomposition}: $K$ is the solution to the minimization problem
\begin{equation*}
    K = \arg \min_{\hat K} \int \norm{\Phi(x)^T \hat K - \Psi(Lx)^T}_{\ell^2}^2 \dd \rho_x(x),
\end{equation*}
i.e., the best linear approximation from $\Phi(x)$ to $\Psi(Lx)$ in the average squared $2$-norm over the distribution of potential inputs. The true distribution $\rho_x$ is inaccessible, so we replace it with the empirical distribution $\hat\rho_x = \frac{1}{n} \sum^n_{i=1} \delta_{x_i}$, which converges weakly almost surely to $\rho_x$ in the large-data limit. Writing $X$ and $Y$ for the data matrices with rows $\Phi(x_i)^T$ and $\Psi(y_i)^T$, the resulting least-squares problem has a unique solution (provided $X$ has full column rank, which requires $n \geq N$):
\begin{equation}
    K = X^\dagger Y =\big(\underbrace{\tfrac{1}{n}X^T X}_{G}\big)^{-1}\,\underbrace{\tfrac{1}{n}X^T Y}_{A},
\end{equation}
where the entries of the Gram matrix $G$ and the cross matrix $A$ are empirical inner products of observables; see Appendix \ref{sec:minimiser} for the full derivation and the interpretation of these matrices. The complete algorithm is summarized in Algorithm \ref{alg:sections}.

Computing a finite section amounts to linear regression between representations, so regression matrices from embeddings can be viewed as finite sections of underlying operators. PRONE rectangular sections of Koopman operators \citep{colbrook2026prone} already generalize DMD, EDMD \citep{colbrook2023multiversedynamicmodedecomposition}, Koopman Regression, and SINDy \citep{brunton2016discovering}. We show that methods from representational similarity analysis (Section \ref{sec:RSA}) and embedding-based regression methods such as DisCoCat \citep{coecke2010mathematical, lo2025discoclip} can likewise be interpreted as finite-section methods for composition operators.

% Calculation of a finite section in this context amounts to a linear regression between representations, allowing regression matrices arising elsewhere on embeddings — RSA methods (Section \ref{sec:RSA}), DisCoCat \citep{coecke2010mathematical, lo2025discoclip}to be viewed as finite sections of underlying operators.
% % Cut for space, kept as a comment:
% % PRONE rectangular sections of Koopman operators \citep{colbrook2026prone} already generalize methods like DMD, EDMD \citep{colbrook2023multiversedynamicmodedecomposition}, Koopman Regression and SINDy \citep{brunton2016discovering}.
% \red{[Compressed this paragraph and cut the PRONE/DMD/SINDy sentence (kept as a source comment).]}

\subsection{Singular Values}
\label{sec:svals}
Suppose we have access to the true finite section $K = \mc{P}_{V_{\Phi}} \CL \mc{P}_{V_\Psi}^*$, where $\Phi$ and $\Psi$ are now orthonormal bases of their respective subspaces. Since $\CL$ is an isometry, its infinite matrix representation with respect to orthonormal bases has mutually orthogonal columns, something the finite sections in general do not. However, we can still perform a Singular Value Decomposition (SVD) on $K$ and interpret the results in an argument similar to performing a CS decomposition on a unitary matrix. Writing the domain and codomain of $\CL$ in the form $\CL: V_{\Psi} \oplus V_{\Psi}^\perp \to V_{\Phi} \oplus V_{\Phi}^\perp$, we obtain a block structure, where $K$ is the finite section obtained previously:
\begin{equation}
    Q = \begin{pmatrix}
        K & K_{12} \\
        K_{21} & K_{22}
    \end{pmatrix}.
\end{equation}
%\red{[Renamed from $V$ to $Q$: $V$ clashed with the subspaces $V_\Phi, V_\Psi$ and with the SVD factor below (itself renamed $V_1$).]}
Since $\CL$ is an isometry, we have $\CL^* \CL = I$, i.e., $Q^T Q = I$ in matrix form (note that $QQ^T \neq I$ in general). Reading off the top-left block of this identity gives $K^T K + K_{21}^T K_{21} = I$. It is then clear that $0 \preceq K^T K \preceq I$ and thus the singular values of $K$ lie in the range $[0,1]$. If we let $U_1 C V_1^T$ be the SVD of $K$:
\begin{equation}
    K_{21}^T K_{21} = V_1 (I - C^2) V_1^T := V_1 S^2 V_1^T,
\end{equation}
and thus if we let $\tilde{\Phi}$ be any orthonormal basis of $V_{\Phi}^\perp$, there exists a matrix $U_2$ (which is infinitely tall but finitely wide) such that $K_{21} = U_2 S V_1^T$. Here $C$ and $S$ play the role of the cosine and sine matrices in the CS decomposition. Using this matrix, we can decompose the action of $\CL$ on the subspace of inputs $V_{\Psi}$ into two parts as follows:
\begin{equation}
\label{eq:decomposition_CL}
    \CL (g^T \Psi) = g^T V_1(C^T U_1^T \Phi + S U_2^T \tilde{\Phi}).
\end{equation}
The first term denotes the part of $\CL$ that maps the span of $\Psi$ perfectly into the span of $\Phi$ and the second term denotes the part of $\CL$ which is lost after projecting onto the span of $\Phi$. The optimal choice of bases $\Phi$ and $\Psi$ would be one which minimizes the second term in \eqref{eq:decomposition_CL}, i.e., which makes the singular values in $C$ as close to $1$ as possible.
%  \red{[The block-structure figure was commented out, leaving a broken reference; it is restored in Appendix \ref{sec:minimiser} to keep it out of the 9-page count.]}

In the general case where $\Phi$ and $\Psi$ are just linearly independent and not orthonormal with respect to the prompt and response distributions, if we let $G_\Phi$ and $G_\Psi$ be the Gram matrices $(G_\Phi)_{ij} = \ip{\phi_i}{\phi_j}_{\rho_x}$ and $(G_\Psi)_{ij} = \ip{\psi_i}{\psi_j}_{\rho_y}$, then $K$ (now the coefficient matrix $G_\Phi^{-1}\mc{P}_{V_\Phi}\CL\mc{P}^*_{V_\Psi}$ of the section, per the convention of Section \ref{sec:fs-cco}) satisfies $K^T G_\Phi K \preceq G_\Psi$ and, with $K_{21}$ still collecting the coefficients of the leakage (with respect to an arbitrary orthonormal basis of $V_\Phi^\perp$),
\begin{equation*}
    K^T G_\Phi K + K_{21}^T K_{21} = G_{\Psi}.
\end{equation*}
% Importantly, the singular values are no longer constrained to be less than $1$, though it is worth noting that if one rescales $K$ by the Gram matrices into $G_{\Phi}^{\frac 12} K G_{\Psi}^{-\frac 12}$, then this new matrix has singular values in the range $[0,1]$. However it remains true that the larger singular values are of most interest.
Importantly, the singular values are no longer constrained to be less than $1$, however it remains true that the larger singular values are of most interest.

\begin{remark}
\label{wachter}
    % It is well known that for $n \times n$ Haar-distributed orthogonal matrices, the top-left $m \times m$ compression matrix (for $m \leq \frac{n}{2}$) has singular values that are distributed according to the $\beta=1$ Jacobi ensemble \citep{edelman2008betajacobi}. In the limit as $n \to \infty$ with $\frac mn \to c$ for a fixed constant $c \in [0,\frac 12]$, the squares of the empirical singular values converge weakly almost surely to the Wachter law $p_c$, whose density is given by
    % \begin{equation}
    % p_c(\lambda)
    % =
    % \frac{1}{2\pi c}
    % \frac{\sqrt{4c(1-c)-\lambda}}
    % {\sqrt{\lambda}(1-\lambda)},
    % \qquad
    % 0<\lambda<4c(1-c).
    % \end{equation} 
    % % Cut for space, kept as a comment:
    % % (In the case where $c > \frac 12$ the distribution has an atom at $1$, corresponding to the singular values which are trivially forced to be $1$).
    % In particular, the singular values have the density
    How much larger should the singular values be? To answer this we consider what happens when the data $X$ and $Y$ is random. Given two Wishart matrices $W_1 \sim \mc{W}_N(n)$ and $W_2 \sim \mc{W}_N(M)$ (recall that if $X$ is an $m \times n$ matrix of i.i.d. standard Gaussians then $\frac 1n X^T X$ has the Wishart distribution denoted $\mc{W}_m(n)$), then the eigenvalue distribution of $\frac{n}{M} W_1^{-1} W_2$ as $\frac NM \to \alpha$ and $\frac{N}{n} \to \beta$ where $\alpha > 0$ and $\beta \in (0,1)$ is given by the density \citep{bai2010spectral}
    \begin{align*}
        p(x) = (1-\beta) \frac{\sqrt{(\lambda_+-x)(x-\lambda_-)}}{2 \pi x (\alpha + \beta x)} \quad x \in [\lambda_-,\lambda_+] \quad \lambda_\pm = \frac{(1\pm h)^2}{(1-\beta)^2} \quad h = \sqrt{\alpha + \beta - \alpha \beta},
    \end{align*}
    along with an atom of strength $\max(0, 1 - \frac{1}{\alpha})$ at $0$.
    One can show that the squared singular values of $X^\dagger Y$ for $X$ and $Y$ being random $n \times N$ and $n \times M$ i.i.d standard Gaussians has the same distribution as $W_1^{-1} W_2$. In the case that $N = M$, the singular values are distributed according to
    \begin{align}
    \label{eq:fdist}
        p(x) = \frac{\sqrt{4\beta-x^{2}(1-\beta)^{2}}}{\pi(1+x^{2})\beta} \qquad x \in [0,b] \qquad b = \frac{2\sqrt{\beta}}{1-\beta}.
    \end{align}
    Although the matrices in practice encountered do not have i.i.d entries, this still provides a null test for measuring the significance of the singular values. We will see an example of this test in Section \ref{sec:comparing} with finite section matrices obtained in practice.
\end{remark}
\subsection{Example: Comparing Model Tasks with Singular Values}
\label{sec:comparing}
The accuracy with which a finite section can capture the spectrum of the LLM operator depends largely on the task it is performing. It is intuitive that complex, open-ended, multi-step tasks are in some way more ``non-linear'', or harder to analyze, and therefore less amenable to approximation via finite sections, but this intuition is qualitative, and the underlying task distributions of Section \ref{sec:cds} are inaccessible. How well a task is captured by a finite section $K$ makes the comparison concrete: as in Section \ref{sec:RSA}, spectral pollution/leakage carries useful comparative information, and we apply the methodology of Section \ref{sec:fs-cco} to make comparative inferences about underlying task distributions.
%  Preprocessing like whitening or dimensionality reduction corresponds to a different choice of observable dictionary. 
From the Dolly-15k instruction corpus \citep{DatabricksBlog2023DollyV2} we consider pools of $220$ prompts in three increasingly open-ended tasks: summarization, brainstorming and creative writing, and construct finite section approximations of the underlying operators induced by these input task distributions $\rho_i$ and various models using paired prompt--response data as in Section \ref{sec:fs-cco}. Both prompts and responses are embedded by the same model, BGE-base $:= \Phi_0$ ($d = 768$) \citep{bge_embedding}. Note here that this doesn't mean that $\Phi = \Psi = \Phi_0$: after mean-centering the embeddings, we reduce dimensionality to $r = 25 \approx n/8$ to avoid rank-deficiency, taking as dictionaries the top-$r$ principal components of each side, $\Phi = W_x^T\Phi_0$ and $\Psi = W_y^T\Phi_0$, where $W_x, W_y \in \R^{d \times r}$ are the loading matrices of the $r$-dimensional principal subspaces $V_x, V_y$ defined by the data. Further details can be found in Appendix \ref{sec:comparison-details}.
\begin{figure}
    \centering
    \includegraphics[width=0.72\linewidth]{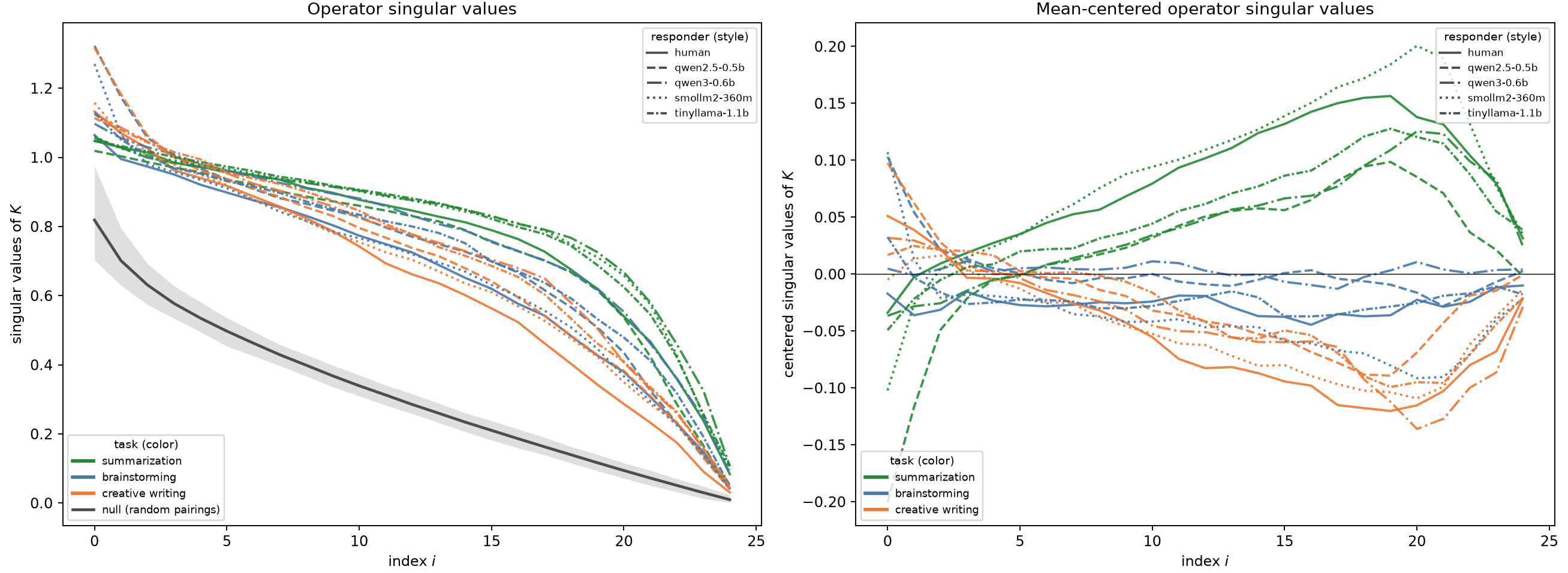}
    \caption{(A) Singular value spectra of the finite sections for each task and responder; (B) singular value spectra after subtracting the mean spectrum across all sections. \vspace{-24pt}}
    \label{fig:task-spectra}
\end{figure}
It is clear from Figure \ref{fig:task-spectra}(A) that the singular values captured by the finite sections of the various LLM operators are non-zero, which validates the intuition that text embedding models serve as good dictionaries when taking sections of LLM operators $\CL$. The leading empirical values  exceed $1$ (Table \ref{tab:sv-stats-category}); this does not contradict Section \ref{sec:svals}, since the $[0,1]$ bound applies to dictionaries orthonormal with respect to the underlying distributions, whereas our principal-component dictionaries are orthogonal but not normalized against $\hat\rho_x$-whitening by the Gram matrices \eqref{proj-whitened} restores the bound: see \ref{whitened-experiment}.

Within a task, the different sections have similar singular values, as seen in Figure \ref{fig:task-spectra}(B). Between tasks, open-endedness corresponds to faster spectral decay / greater spectral leakage. In relation to \eqref{eq:decomposition_CL}, less of the subspace $V_\Psi \subset \mc{H}_y$ is mapped into $V_\Phi \subset \mc{H}_x$ as the open-endedness of the task increases; this could correspond to the variance, or support, of the response distribution $\rho_y$ being larger in more open-ended tasks.

Lastly, in Figure \ref{fig:task-spectra}(A), we find the singular values of the operator sections exceed those of a null operator constructed from random permutations of prompt-response pairs. Equation \ref{eq:fdist} also gives the upper support of the singular values for i.i.d. Gaussian matrices $X$ and $Y$. Although random permutation matrices do not satisfy this assumption, the largest singular value matches closely to \eqref{eq:fdist} when $\beta = \frac 18$, which we show in \ref{remark-appendix}.
\vspace{-8pt}

\section{Discussion and Future Work}

\textbf{Stochasticity of LLMs.} An LLM is stochastic, mapping prompts to distributions rather than points. \citep{colbrook2024beyond} introduce the Koopman operator for random systems. Composition operators in the random case is left as future work.

\noindent \textbf{Analysis of Interacting Agents.} Since composition operators convert composition of maps into operator multiplication (Section \ref{sec:intro}), this is a natural setting for linearizing an agent that has a specific task and expected response format. Rectangular free convolution calculates the singular value spectrum of a product of matrices from their individual spectra; we foresee using this to predict the spectra, and hence the downstream behavior, of finite compositions of agents.

% \textbf{Word Matrix Representations.} Linguistic tools like DisCoCat \citep{coecke2010mathematical} construct matrix representations of words, e.g., encoding adjectives as matrices that act on nouns. The ``compositional gap'' is a problem at the syntax--semantics interface where the composition of word-level representations fails to track the meaning of the composed phrase; finite sections of composition operators may offer a way to quantify this gap.

\newpage

\bibliography{references}

\newpage

\appendix

\section{Derivation of the Finite Section Minimization}
\label{sec:minimiser}

Fix an arbitrary $g \in V_\Psi$. Then there is a vector $a \in \R^M$ such that $g = a^T \Psi$ and the action of $\CL$ on $g$ can be written as
\begin{equation}
    \CL g(x) = g(Lx) = \Psi(Lx)^T a = \Phi(x)^T K a + \underbrace{\big(\Psi(Lx)^T - \Phi(x)^T K\big)}_{r(x)^T}\, a,
\end{equation}
where we have defined the remainder $r(x)$. The ``best'' $K$ here minimizes the residual term $r(x)^T a$ over all input observables up to scaling, i.e., over all $a$ with $\norm{a}_2 = 1$. The choice to only minimize over $a$ having unit $2$-norm is somewhat arbitrary; one can minimize instead over all $a$ such that $\norm{D a} = 1$ where $D$ is a positive-definite matrix, however one can show that the solution of the minimization problem that follows (i.e. without regularization) does not depend on $D$, so we will fix it to be the identity matrix. Thus, $K$ is defined as
\begin{equation*}
    K = \arg\min_{\hat{K}} \int_{\Sigma^*} \max_{\norm{a}_2 = 1} |r(x)^T a|^2 \dd\rho_x(x).
\end{equation*}
This inner term is maximized when $a$ is in the direction of $r(x)$. Setting $a = \frac{r(x)}{\norm{r(x)}_2}$ yields
\begin{equation*}
    K = \arg \min_{\hat K} \int \norm{\Phi(x)^T \hat K - \Psi(Lx)^T}_{\ell^2}^2 \dd \rho_x(x).
\end{equation*}
This expression has the interpretation of the minimization of the average squared $2$-norm residual over the distribution of potential inputs. Note that this expression contains Equation \eqref{eq:linear_approx} and thus can be interpreted as finding the best linear approximation from $\Phi(x)$ to $\Psi(Lx)$ under a specific norm.

The true distribution $\rho_x$ is inaccessible, but we can use a data-driven approximation, based around the empirical distribution $\hat\rho_x = \frac{1}{n} \sum^n_{i=1} \delta_{x_i}$, which converges weakly almost surely to $\rho_x$ in the large-data limit. Hence we can approximate the integral appearing in the minimization problem as
\begin{align*}
    \int \norm{\Phi(x)^T \hat K - \Psi(Lx)^T}_{\ell^2}^2 \dd \rho_x(x) & \approx \int \norm{\Phi(x)^T \hat{K} - \Psi(Lx)^T}_{\ell^2}^2 \dd \hat\rho_x(x) \\
    &= \frac{1}{n} \sum_{k=1}^n \norm{\Phi(x_k)^T \hat K - \Psi(y_k)^T}_{\ell^2}^2 = \frac{1}{n} \norm{X \hat K - Y}_{F}^2,
\end{align*}
where the data matrices $X$ and $Y$ have rows $\Phi(x_i)^T$ and $\Psi(y_i)^T$ respectively. By replacing the integral with the discrete version, the minimization problem has a unique solution in terms of the pseudo-inverse of $X$ or the Gram matrix (provided $X$ has full column rank, which requires $n \geq N$):
\begin{equation}
    K = X^\dagger Y =\big(\underbrace{\tfrac{1}{n}X^T X}_{G}\big)^{-1}\,\underbrace{\tfrac{1}{n}X^T Y}_{A},
\end{equation}
where we have defined the matrices $A$ and $G$. Below we quickly discuss the interpretation of these two matrices.

The data matrices $X, Y$ can be reinterpreted as columns of the individual observables $\phi_i$ and $\psi_i$ sampled at the points $x_i$. This is denoted $\boldsymbol{\phi}_i = [\phi_i(x_1),...,\phi_i(x_n)]^T \in \bb{R}^{n}$, and equivalently for $\boldsymbol{\psi}_i$ and $\CL(\boldsymbol{\psi}_i)$.

\begin{equation}
    X = \begin{bmatrix}
-\,\Phi(x_1)^T\,- \\
\vdots \\
-\,\Phi(x_n)^T\,-
\end{bmatrix} =
\begin{bmatrix}
    \vert &   & \vert \\
    \boldsymbol{\phi}_1  & \cdots & \boldsymbol{\phi}_N \\
    \vert &   & \vert
\end{bmatrix}
\end{equation}

\begin{equation}
Y = \begin{bmatrix}
-\,\Psi(y_1)^T\,- \\
\vdots \\
-\,\Psi(y_n)^T\,-
\end{bmatrix} =
\begin{bmatrix}
-\,(\Psi \circ L(x_1))^T\,- \\
\vdots \\
-\,(\Psi \circ L(x_n))^T\,-
\end{bmatrix} =
\begin{bmatrix}
    \vert &   & \vert \\
    \CL(\boldsymbol{\psi}_1)  & \cdots & \CL(\boldsymbol{\psi}_M) \\
    \vert &   & \vert
\end{bmatrix}
\end{equation}
where $\CL(\boldsymbol{\psi}_i)$ is defined as the vector with entries $(\CL\psi_i)(x_j)$. From this, we see that the finite section $K$ is the product of $G^{-1}$ and $A$, where the entries of $G$ and $A$ are given by
\begin{equation}
    G_{jk} = \tfrac{1}{n}\boldsymbol{\phi}_j^T \boldsymbol{\phi}_k = \tfrac{1}{n}\sum^n_{i=1}\phi_j(x_i) \phi_k(x_i) = \langle\phi_j,\phi_k \rangle_{\hat\rho_x} \qquad A_{jk} =
    \ip{\phi_j}{\CL\psi_k}_{\hat\rho_x}.
\end{equation}
Thus, the entries of $A$ and $G$ are inner products of the vectors formed by observables applied to the datapoints $(x_i, y_i)$.

\subsection{Algorithm for Calculating $K$}

\begin{algorithm}
    \caption{Computing a finite section of the composition operator $\CL$ (without whitening)}
    \label{alg:sections}
    \DontPrintSemicolon
    \KwIn{Dictionaries $\Phi : \mc{X} \to \R^N$ and $\Psi : \mc{Y} \to \R^M$ and paired data $(x_i, y_i)_{i=1}^{n}$ with $y_i = L x_i$}
    \KwOut{Approximate finite section $K \in \R^{N \times M}$}
    \qquad $X \gets \begin{bmatrix} \Phi(x_1) & \cdots & \Phi(x_n) \end{bmatrix}^T \in \R^{n \times N}$\;
    
    \qquad $Y \gets \begin{bmatrix} \Psi(y_1) & \cdots & \Psi(y_n) \end{bmatrix}^T \in \R^{n \times M}$\;
    
    \qquad $K \gets X^\dagger Y$\;
    \BlankLine
    \KwRet{$K$}
\end{algorithm}

% \begin{figure}
%     \centering
%     \begin{tikzpicture}[baseline=(current bounding box.center)]
%     % Shading
%     \fill[blue!40] (0,4) rectangle (2,6);
%     \fill[red!40] (0,0) rectangle (2,4);

%     % Main square
%     \draw (0,0) rectangle (6,6);

%     % Block divisions
%     \draw (2,0) -- (2,6);
%     \draw (0,4) -- (6,4);

%     % Block labels
%     \node at (1,5) {$K$};
%     \node at (4,5) {$K_{12}$};
%     \node at (1,2) {$K_{21}$};
%     \node at (4,2) {$K_{22}$};

%     % Top braces
%     \draw[decorate,decoration={brace,amplitude=8pt}]
%         (0,6.15) -- (2,6.15)
%         node[midway,above=8pt] {$V_{\Psi}$};

%     \draw[decorate,decoration={brace,amplitude=8pt}]
%         (2,6.15) -- (6,6.15)
%         node[midway,above=8pt] {$V_{\Psi}^{\perp}$};

%     % Left braces
%     \draw[decorate,decoration={brace,amplitude=8pt}]
%         (-0.15,4) -- (-0.15,6)
%         node[midway,left=8pt] {$V_{\Phi}$};

%     \draw[decorate,decoration={brace,amplitude=8pt}]
%         (-0.15,0) -- (-0.15,4)
%         node[midway,left=8pt] {$V_{\Phi}^{\perp}$};
%     \end{tikzpicture}
%     \caption{Matrix representation of $\CL$ with respect to the finite-dimensional subspaces $V_\Psi$ and $V_\Phi$, where $K$ (blue) denotes the finite section which can be computed as in Section \ref{sec:fs-cco} and $K_{21}$ (red) denotes the components of $V_\Psi$ which are mapped out of the span of $V_\Phi$ and cause spectral leakage.}
%     \label{fig:matrix_composition}
% \end{figure}

\section{Further Details of Example \ref{sec:comparing}}
\label{sec:comparison-details}

Anonymized GitHub Repository: \\ \url{https://anonymous.4open.science/r/task-sections-318C}

Our experiment uses 220 prompts from 3 categories: summarization, brainstorming and creative writing from the Dolly-15k instruction corpus \citep{DatabricksBlog2023DollyV2}. We consider human responses to the prompts, and generated responses from four other models: Qwen3-0.6B \citep{yang2025qwen3}, SmolLM2-360M \citep{allal2025smollm2}, Qwen2.5-0.5B \citep{qwen2024qwen25}, TinyLlama-1.1B \citep{zhang2024tinyllama}.

In Section \ref{sec:comparing}, we use the BGE-base model \citep{bge_embedding} ($d=768$). In this section we present similar results when we whiten the BGE embeddings which is equivalent to CCA, and results for a smaller model, MiniLM \citep{wang2020minilm} ($d=384$).

\begin{table}[H]
  \centering
  \caption{Summary statistics of the operator singular values per task category, mean $\pm$ sd across the five responders (human, Qwen3-0.6B, SmolLM2-360M, Qwen2.5-0.5B, TinyLlama-1.1B). Each operator is the rank-truncated section $K$ for $(n{=}200,\, r{=}25)$; every statistic is averaged over 20 stimulus subsamples. $s_1$: largest singular value; $\bar{s}$: mean singular
  value; $\overline{s^2}$: mean squared singular value.}
  \label{tab:sv-stats-category}
  \begin{tabular}{lccc}
    \toprule
    Task & $s_1$ & $\bar{s}$ & $\overline{s^2}$ \\
    \midrule
    Summarization    & $1.046 \pm 0.014$ & $0.783 \pm 0.025$ & $0.673 \pm 0.034$  \\
    Brainstorming    & $1.177 \pm 0.101$ & $0.710 \pm 0.038$ & $0.584 \pm 0.051$  \\
    Creative writing & $1.169 \pm 0.076$ & $0.686 \pm 0.033$ & $0.561 \pm 0.041$ \\
    \midrule
    Null (random pairings) & $0.819 \pm 0.059$ & $0.317 \pm 0.005$ & $0.148 \pm 0.008$ \\
    \bottomrule
  \end{tabular}
\end{table}

\subsection{MiniLM Results}

\begin{table}[H]
  \centering
  \caption{Summary statistics on the MiniLM dictionary (all-MiniLM-L6-v2, 384-d), unwhitened rank-truncated sections. Identical setup to Table \ref{tab:sv-stats-category}.}
  \label{tab:sv-stats-minilm}
  \begin{tabular}{lccc}
    \toprule
    Task & $s_1$ & $\bar{s}$ & $\overline{s^2}$ \\
    \midrule
    Summarization    & $1.039 \pm 0.009$ & $0.782 \pm 0.024$ & $0.663 \pm 0.035$ \\
    Brainstorming    & $1.163 \pm 0.101$ & $0.707 \pm 0.049$ & $0.581 \pm 0.064$ \\
    Creative writing & $1.141 \pm 0.073$ & $0.706 \pm 0.038$ & $0.577 \pm 0.050$ \\
    \midrule
    Null (random pairings) & $0.785 \pm 0.069$ & $0.317 \pm 0.009$ & $0.147 \pm 0.011$ \\
    \bottomrule
  \end{tabular}
\end{table}

\begin{figure}[H]
    \centering
    \includegraphics[width=1\linewidth]{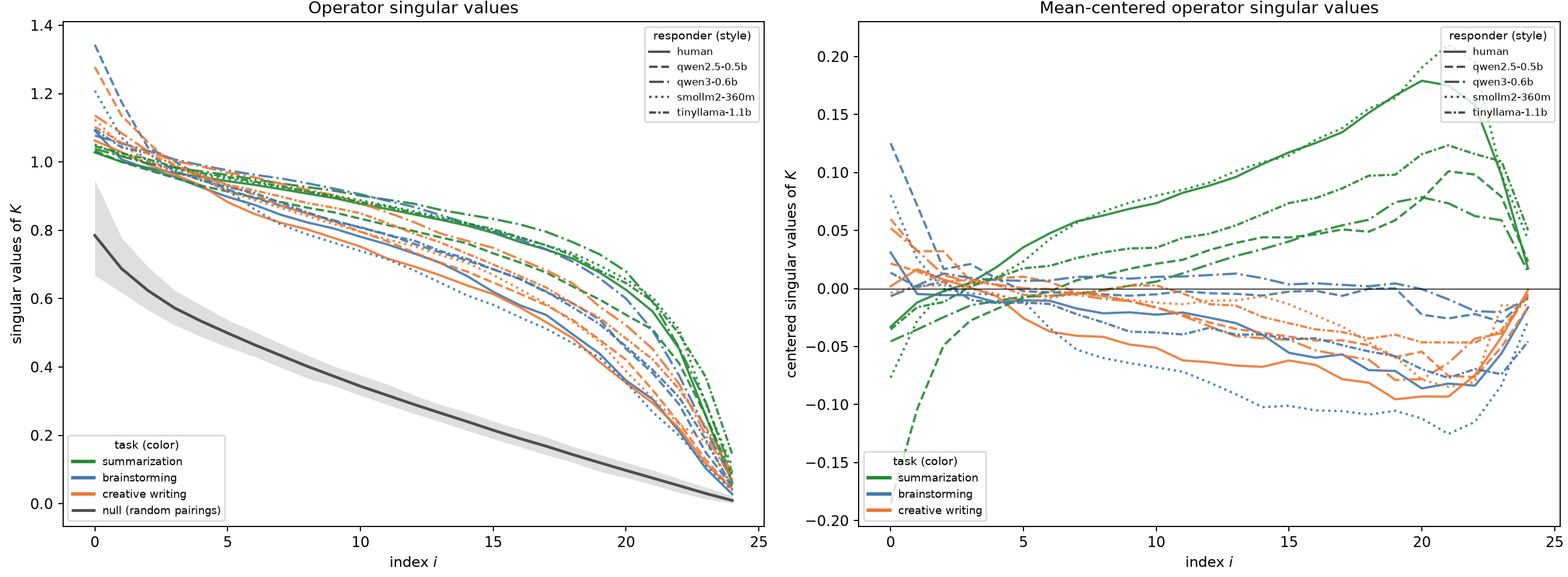}
    \caption{MiniLM results, analogous to Figure \ref{fig:task-spectra}.}
    \label{fig:minilm-dual}
\end{figure}

\subsection{Whitened BGE Results}
\label{whitened-experiment}

\begin{table}[H]
  \centering
  \caption{Summary statistics on whitened rank-truncated sections (canonical correlations) of BGE embedding model. Analagous to previous experiments.}
  \label{tab:sv-stats-category-whitened}
  \begin{tabular}{lccc}
    \toprule
    Task & $s_1$ & $\bar{s}$ & $\overline{s^2}$ \\
    \midrule
    Summarization    & $0.979 \pm 0.005$ & $0.787 \pm 0.026$ & $0.677 \pm 0.034$ \\
    Brainstorming    & $0.964 \pm 0.010$ & $0.706 \pm 0.038$ & $0.562 \pm 0.050$ \\
    Creative writing & $0.959 \pm 0.012$ & $0.680 \pm 0.035$ & $0.531 \pm 0.045$ \\
    \midrule
    Null (random pairings) & $0.635 \pm 0.023$ & $0.303 \pm 0.009$ & $0.126 \pm 0.007$ \\
    \bottomrule
  \end{tabular}
\end{table}

\begin{figure}[H]
    \centering
    \includegraphics[width=1\linewidth]{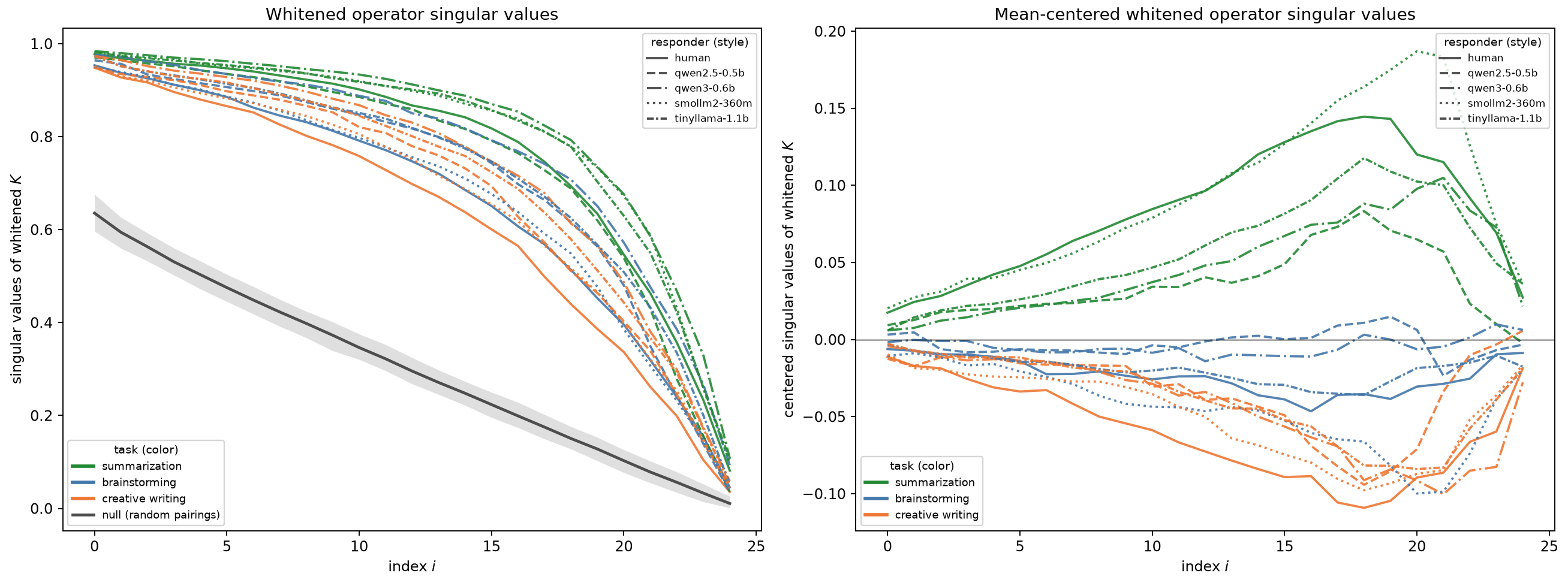}
    \caption{BGE Whitened results, analogous to Figure \ref{fig:task-spectra}.}
    \label{fig:placeholder}
\end{figure}
\subsection{Relation to Remark \ref{wachter}}
\label{remark-appendix}

Remark \ref{wachter} states that the singular value spectrum of random (gaussian i.i.d.) data matrices $X, Y$ follows distribution $p$ defined \eqref{eq:fdist}, and suggests that this distribution can be used to identify when the singular value spectrum of our operator is more than random. In our experiments we compare against an empirical null singular value spectrum is obtained using neither gaussian or i.i.d. data so it might seem to be unrelated to this distribution. We compare the quantile plot of $p$ against the null distributions in the unwhitened experiments on the BGE-base and MiniLM models. Note that we do not expect the experiment with whitened representations to follow this distribution; one can show that the squared singular values should instead asymptotically follow the Jacobi law \citep{wachter1980strong}.

\begin{figure}
    \centering
    \includegraphics[width=1\linewidth]{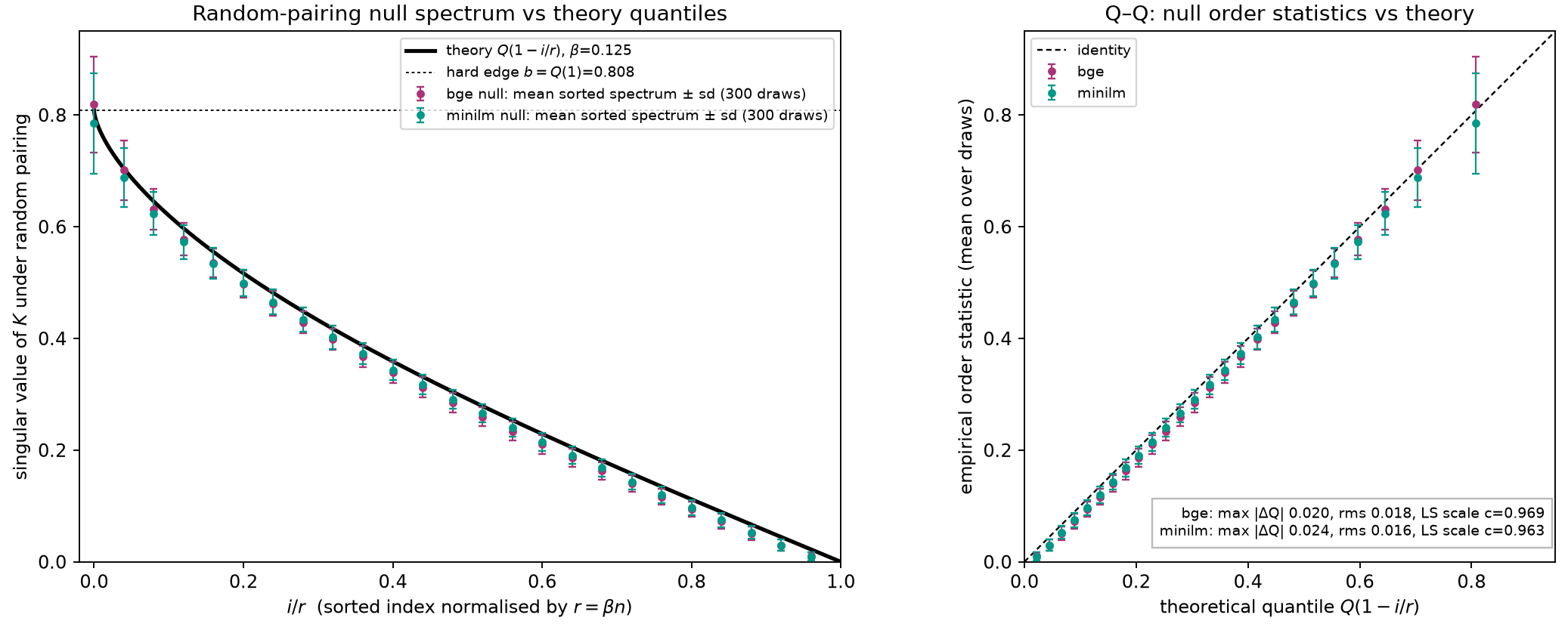}
    \caption{(A) Empirical null spectra vs Theoretical distribution $p$ in \eqref{eq:fdist} for $\beta=0.125$, (B) Q-Q comparison of the empirical and theoretical singular value distributions.}
    \label{fig:placeholder}
\end{figure}

\end{document}